\documentclass{article}

\usepackage[final]{nips_2018}

\usepackage[utf8]{inputenc} 
\usepackage[T1]{fontenc}    
\usepackage{hyperref}       
\usepackage{url}            
\usepackage{booktabs}       
\usepackage{amsfonts}       
\usepackage{nicefrac}       
\usepackage{microtype}      
\usepackage{amsmath, amssymb}

\usepackage{natbib}
\usepackage{graphicx}
\usepackage{color, soul}

\newcommand{\xx}{\mathbf{x}}
\newcommand{\dd}{\mathbf{d}}
\newcommand{\zz}{\mathbf{z}}
\newcommand{\expect}[2]{\mathbb{E}_{#1}[#2]}
\usepackage{multirow}

\title{Robustness against the channel effect in pathological voice detection }

\author{
 Yi-Te Hsu$^{1,3,4}$, Zining Zhu$^{3,4}$, Chi-Te Wang$^6$, Shih-Hau Fang$^7$, Frank Rudzicz$^{2,3,4,5}$, Yu Tsao$^{1}$\\
  $^{1}$: Academia Sinica 
  $^{2}$: International Centre for Surgical Safety, St. Michael's Hospital \\  
  $^{3}$: Department of Computer Science, University of Toronto\\
  $^{4}$: Vector Institute for Artificial Intelligence
  $^{5}$: Surgical Safety Technologies Inc.\\
  $^{6}$: Far Eastern Memorial Hospital;
  $^{7}$: Yuan Ze University\\
  \texttt{b01901112@ntu.edu.tw, zining@cs.toronto.edu, drwangct@gmail.com,}\\
  \texttt{shfang@saturn.yzu.edu.tw, frank@cs.toronto.edu, yu.tsao@citi.sinica.edu.tw}}

\begin{document}

\maketitle
\begin{abstract}
Many people are suffering from voice disorders, which can adversely affect the quality of their lives. 
In response, some researchers have proposed algorithms for automatic assessment of these disorders, based on voice signals. However, these signals can be sensitive to the recording devices. Indeed, the channel effect is a pervasive problem in machine learning for healthcare.
In this study, we propose a detection system for pathological voice, which is robust against the channel effect. This system is based on a bidirectional LSTM network. To increase the performance robustness against channel mismatch, we integrate domain adversarial training (DAT) to eliminate the differences between the devices. 
When we train on data recorded on a high-quality microphone and evaluate on smartphone data without labels, our robust detection system increases the PR-AUC from $0.8448$ to $0.9455$ (and $0.9522$ with target sample labels).
To the best of our knowledge, this is the first study applying unsupervised domain adaptation to pathological voice detection. 
Notably, our system does not need target device sample labels, which allows for generalization to many new devices.

\end{abstract}

\section{Introduction}
\label{introduction}
\vspace{-3pt}
\citet{populationPatho} reported that around $7.6\%$ people in the United States suffer from voice pathology.
They have difficulties communicating with others, which can substantially reduce their quality of life. Most importantly, such diseases usually get worse if the situation is not treated at an early stage, resulting in substantial costs for individuals and for public insurances, depending on jurisdiction.
Unfortunately, two main factors make detecting voice pathology difficult. First, it can be hard for individuals to notice their own degenerative voice problems in subtle stages, which are more evident to speech-language pathologists. Second, the pathologists may have limited availabilities due to ongoing pressures on the profession. Consequently, assessment is currently not optimal.

To enable early automatic diagnosis of pathological speech, 
some studies used traditional speech methods, such as time-frequency approaches \citep{pathoMethod1}, Mel frequency cepstral coefficients (MFCCs) \citep{pathoMethod5}, Gaussian mixture models (GMMs) \citep{pathoMethod2}, hidden Markov models \citep{pathoMethod3} and wavelet coefficients \citep{pathoMethod4} to design the model for differentiating between normal and pathological voice samples.
\citet{canInt16} confirmed the feasibility of using an automatic speech recognition system for voice and speech disorders analysis and assessment. \citet{wuInt18} proposed to use a convolutional deep belief network (CDBN) to pre-train the weights of a CNN (convolutional neural network) model for pathological voice detection; with the pre-training process, the system can yield satisfactory performance even only a small amount of training data is available.
\citet{pathoMethod6} reached the 98\% accuracy on the Massachusetts Eye and Ear Infirmary (MEEI) voice pathology database \citep{pathoMethod1} by combining two classifiers.
\citet{Fangdetection} first time used deep multilayer perceptrons (MLPs), to identify pathological voice and achieved 99\% accuracy on the same dataset. These studies have shown possibilities to discriminate pathological voice through speech processing techniques.

However, these methods have yet to be made widely accessible to the public. 
We wish to balance the acuity of high-quality recordings with the availability of modern mobile devices, and to use models trained on either one.
There are two main challenges -- the first is the relative sparsity of device-specific data, and the second is the wide array of potential devices.

Along similar lines, \citep{cnntransfer} used a fine-tuned CNN model to transfer voice samples from one device to a target device. However, this approach relies on the labeled samples from target device, which is usually costly to acquire in reality. 
Therefore, it is necessary to design an unsupervised architecture which can adapt to target device with unlabeled target samples.
In this work, we design a system which can minimize the channel effects between the recording devices. Our framework addresses the problem of few and unlabeled target samples with unsupervised learning.


\section{Methodology}
\label{methodology}
\vspace{-3pt}
    \begin{figure}[ht]
        \centering
        \includegraphics[height=4.5cm]{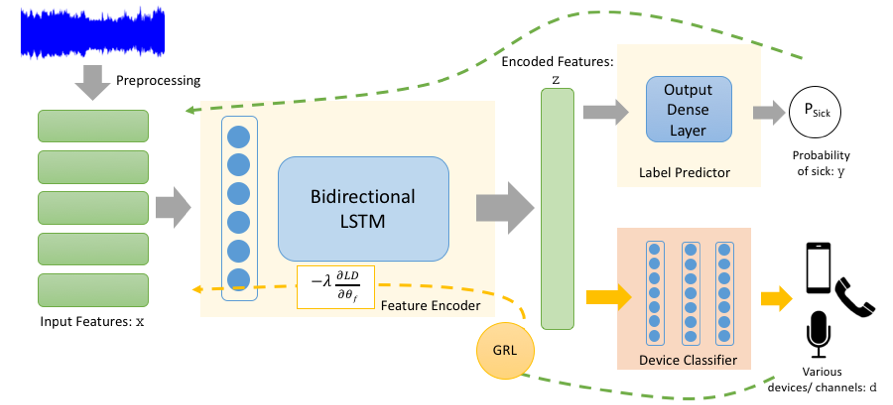}
        \caption{The proposed robust pathological voice detection system. Data preprocessing transforms the voice samples to input features $\mathbf{x}$ (MFCCs or Filter banks). The bidirectional long short-term memory (BLSTM) network encodes input features $\mathbf{x}$ to $\mathbf{z}$. The label predictor forwards the encoded features $\mathbf{z}$ and estimates the probability $y$ of pathological voice. 
        The gradient reversal layer (GRL) is inserted before feeding $\zz$ to the device classifier. GRL multiplies the gradient by $-1$ during backward propagation. With GRL, the encoder, BLSTM, learns features without any device information.}
        \label{fig:dannstructure}
    \end{figure}

\subsection{Front-end feature extraction}
\vspace{-5pt}
We extract 26-dimensional MFCCs and 40-dimensional filter banks features, following conventions in acoustic processing literature \citep{Ravikumar2011,Ganchev2005}. To avoid a loss of information in boundary effects, we set half of the window length as the frame-shift.


\vspace{-3pt}
\subsection{Back-end robust pathological voice detection system}
\vspace{-5pt}
In our formulation, each data point is represented by the triplet \{\textit{input utterance}, \textit{device}, \textit{label}\}, denoted by $(\xx, \dd, y)$.  Our full model consists of a bidirectional long short-term memory (BLSTM) encoder, a dense layer label predictor, and a device classifier whose parameters are denoted by $\theta_z, \theta_y, \theta_d$ respectively.  The BLSTM maps the input utterance to an embedding, $f_{\theta_z}:\xx \rightarrow \zz$.  

\vspace{-3pt}
\subsubsection{Detection model}
\vspace{-5pt}
Our detection model contains a BLSTM \citep{Hochreiter1997} that encodes a latent vector $\mathbf{z}$ for each input audio clip $\mathbf{x}$. On top of it, a fully-connected dense layer classifier estimates the probability of the voice being pathological: $\hat{y}=\text{argmax}_y P(y|\zz; \theta_y)$.

\subsubsection{Unsupervised domain adaptation} 
\vspace{-5pt}
In order to generalize the detection model across devices, it is necessary to eliminate effects of channel mismatch caused from different devices. 
To address two main difficulties (i.e., the small number of target samples and the lack of labels), we propose an unsupervised domain adaptation system as following. Similar to \citet{dann}, our model is encouraged to learn a domain-invariant embedding without using target domain labels. The training scheme can be written as:
\vspace{-10pt}

\begin{align}
\label{eq:model-loss-original}
     \max_{\theta_d} & \min_{\theta_z, \theta_y}  \expect{\xx \in \mathcal{D}_s \cup \mathcal{D}_t}{L_y - \lambda L_d} \\
    L_y &= -\log P(y|\xx;\theta_z, \theta_y)\\
    L_d &= -\log P(\dd|\xx;\theta_z, \theta_d)
\end{align}
where $\mathcal{D}_s$ and $\mathcal{D}_t$ denote source and target domain data respectively. The encoder and label predictor minimize the negative log-probability of the label. 
The device classifier maximizes the log-probability of device classifier as usual, while the encoder minimizes the log-probability of device classifier.
Intuitively, the BLSTM encoder tries to learn an embedding that makes label prediction easy but device classification difficult (i.e., by wiping out information related to devices).

As shown in Fig.~\ref{fig:dannstructure}, the BLSTM encodes each input sample into an embedding $\mathbf{z}$. We implement the simultaneously maximize and minimize over the device loss by inserting a gradient reversal layer (GRL). 
The GRL has no effect during the forward pass, but it multiplies the gradient by $-1$ in back propagation.  

In this work, we focus on unsupervised domain adaptation. We modify the formulation, assuming no knowledge about the labels $y$ for the voice sample in the target device $\mathcal{D}_t$, which means no $\mathcal{L}_y$ in the formulation, as following:
\begin{align}
\label{eq:model-loss-modified}
    & \max_{\theta_d} \min_{\theta_z, \theta_y}  \expect{\xx \in \mathcal{D}_s}{L_y - \lambda L_d} + \expect{\xx \in \mathcal{D}_t}{- \lambda L_d} 
\end{align}

In Experiments (\S \ref{sec:experiments}), we show that our robust pathological voice detection system can increase the area under the precision-recall curve (PR-AUC) significantly to solve the channel mismatch between different devices.

\section{Experiments and results}
\label{sec:experiments}
We first compare the two deep learning models, BLSTM and multilayer perceptron (MLP) models, and examine the performance under numerous feature extraction settings. The details of models design are in Appendix \ref{a:model}.
Based on these experiments, we select the best features and model to apply to our unsupervised domain adaptation system.
We also compare our unsupervised domain adaptation system with three baseline results. 
We evaluate our results using the area under the precision-recall curve (PR-AUC), which is particularly suited to classifiers given imbalanced datasets.

\paragraph{{Dataset}}
Each audio sample was collected from the Far Eastern Memorial Hospital (FEMH), as detailed in Appendix \ref{a:datadescription}. There were two voice collection mechanisms, recorded by a microphone and a smartphone.
A vowel sound /:a/ was uttered in each recording. 
The pathological samples are the voice affected by the vocal fold pathologies.
The source domain contained 183 voice samples (133 pathological samples and 50 control samples); we randomly choose 146 samples as the training set, and the other 37 samples as test set. The target domain (smartphone dataset) includes 52 pathological and 20 control samples; here, we use 26 pathological and 10 control samples as the test set. 
Note that the two sets differ in both recording devices and speakers. 

\subsection{MLP versus Bidirectional LSTM}
We use 26-dimension MFCCs and 40-dimension filter banks with 32 ms windows. We also compare the effect of normalizing features over time to select the right model. 
    \begin{table}[h]
		\centering 
		\begin{tabular}{c c c c c} 
			\hline\hline 
			\multirow{2}{*}{Model}&\multicolumn{2}{c}{MFCCs} & \multicolumn{2}{c}{Filter banks}\\
            & Normalized & Non-normalized & Normalized & Non-normalized\\
            \hline
            BLSTM & 0.94150 & 0.9051 & 0.8765 & 0.9478 \\
            MLP & 0.91541 & 0.8693 & 0.8109 & 0.8279 \\
			\hline 
		\end{tabular}
		\caption{PR-AUC scores of BLSTM and MLP frameworks with features: MFCCs, normalized MFCCs, filter banks and normalized filter banks.
		\label{tab:DNNvsBLSTM}}
	\end{table}

\subsection{Analysis of MFCCs and filter banks under various setting}
The experiment compares MFCCs and filter banks across different settings, including window length and  normalization. We use the BLSTM because  Tables~\ref{tab:DNNvsBLSTM} show that it outperforms MLP  in this task.

    \begin{table}[h]
		\centering 
		\begin{tabular}{c c c c c} 
			\hline\hline 
			\multirow{ 2}{*}{Feature} &\multicolumn{2}{c}{32ms (window length)} & \multicolumn{2}{c}{100ms (window length)}\\
            & Normalized & Non-normalized & Normalized & Non-normalized\\
            \hline
            MFCCs & \textbf{0.94150} & 0.9051 & \textbf{0.9350} & 0.9111 \\
            Filter banks & 0.8765 & \textbf{0.9478} & 0.8444 & \textbf{0.9378} \\
			\hline 
		\end{tabular}
		\caption{PR-AUC scores of MFCCs and filter banks with different setting. Each score is the mean of three different runs with distinct random seeds.
		\label{tab:featureSelection}}
	\end{table}
As shown in Table~\ref{tab:featureSelection}, for MFCCs, normalization improves performance; by contrast, for filter banks, non-normalized models performs better. 

\subsection{Domain adaptation}
\label{domain_adaptation}
Previous experiments showed that non-normalized filter banks with 32 ms window length best capture the characteristics of data samples, so we use these as features to evaluate our domain adaptation model. 
We set up the following three baselines.
\paragraph{Baseline 1: No adaptation} Here, we train on the source device and test on the target device. The resulting poor performance demonstrates the channel mismatch effect (i.e., the discrepancy between data recorded from different devices).
\paragraph{Baseline 2: Target domain only} Here, we train and test on different subsets of the small target device dataset. Here, the low model performance reflects the impact of small dataset size.
\paragraph{Baseline 3: Frozen layers} After training the BLSTM classifier on the source device data, we freeze the BLSTM part and fine-tune the dense classifier layer using a subset of the target domain data. The frozen layer models perform better than the other two baselines but its performance is worse than our proposed detection system, which is combining the domain adversarial training (DAT) technique. Training on the source domain apparently overfits, and the decrease in the model's generalizability can not be easily recovered by fine-tuning on limited data on the target domain.


    \begin{table}[h]
		\centering 
		\begin{tabular}{c l c} 
			\hline\hline 
            \hspace{1cm} & \hspace{5em} Method & PR-AUC \\
            \hline
            
            \multirow{4}{*}{Supervised} & Baseline 1: No domain adaptation & 0.8448 \\
            & Baseline 2: Target domain only & 0.8509 \\
            & Baseline 3: Frozen layer fine-tuning & 0.9021 \\
            & Proposed system: DAT with target labels (Equation \ref{eq:model-loss-original})* & 0.9522 \\
            \hline
            Unsupervised & Proposed system: DAT (Equation \ref{eq:model-loss-modified})* & 0.9455 \\
			\hline 
		\end{tabular}
		\caption{PR-AUC results of domain adaptation. * Both proposed detection systems significantly outperform the other three baseline methods (p-value < 0.05 by t-test, while there is no significant difference between these two). 
		\label{tab:transfer_result}}
	\end{table}

\section{Conclusion and discussion}
\label{conclusion}

In this work, we propose a robust pathological voice detection system and evaluate its performance using PR-AUC and examine the performance of MFCCs and filter banks with different settings. To the best of our knowledge, this is the first study that investigates this task based on the unsupervised adaptation approach. Our
BLSTM model achieves a PR-AUC score of $0.94$. Furthermore, to facilitate the deployment of the algorithm to mobile phones, we integrate unsupervised domain adaptation method, which increases the PR-AUC score from $0.84$ to $0.94$ without labeled target device samples. 

\bibliographystyle{plainnat}
\bibliography{nips_2018}

\section{Appendix}
\label{Appendix}

\subsection{Data description}
\label{a:datadescription}

There are two collections, one is the larger dataset, which is recorded by the high-quality microphone; the other one is the smaller dataset, which is recorded by the smartphone (iPhone). 
The sampling rate was 44,100 Hz with a 16-bit resolution, and data were saved in an uncompressed wave format. Within voice samples, a vowel sound /:a/ was recorded. The pathological samples are the voice affected by the vocal fold pathologies.

The source domain (microphone) contained 183 voice samples (133 pathological samples and 50 control samples); we randomly choose 146 samples as the training set, and the other 37 samples as test set. The target domain (smartphone dataset) includes 52 pathological and 20 control samples; here, we use 26 pathological and 10 control samples as the test set.

\subsection{Feature extraction procedures}
\label{a:featureextraction}

\paragraph{Filter banks} can be derived through a series of calculations, pre-emphasis, framing, windowing, Fourier transform, and Mel filtering. These procedures aim to mimic non-linear human ear perception of sound. Filter banks are powerful feature extractors. Therefore, we introduce this approach as one of our feature extraction methods.  

\paragraph{Mel-Frequency Cepstral Coefficients (MFCCs)} are the speech features derived from filter banks by applying a discrete cosine transform (DCT) to remove the correlation between filter bank coefficients. It can thus be regarded as the compressed representation of the filter banks. MFCCs are widely used in speech recognition, speaker recognition, and other speech processing tasks. 

\subsection{Model}
\label{a:model}
For DNN-based multilayer perceptron (MLP) model, we follow the setting in \citet{Fangdetection}. The MLP model has three dense layers, each with 300 nodes and predicts the result by each frame. For the BLSTM classification model, at each step, the model takes as input the concatenation of the eleven neighboring frames. The BLSTM model has a fully connected dense layer with 512 nodes, two bidirectional LSTM layers, each with 512 nodes, and a fully connected output layer to classify two classes. Both models use ReLU as the activation function. 
For the domain classifier in the Fig.~\ref{fig:dannstructure}, the DNN-based devices classifier consists of 3 dense layers, each with 300 nodes and the activation function is ReLU.
We implement all models in TensorFlow \citep{tensorflow}.

\end{document}